\documentclass[preprint]{article}

\usepackage[preprint]{neurips_2025}

\usepackage[utf8]{inputenc}
\usepackage[T1]{fontenc}
\usepackage{hyperref}
\usepackage{url}
\usepackage{booktabs}
\usepackage{amsfonts}
\usepackage{amsmath}
\usepackage{microtype}
\usepackage{xcolor}
\usepackage{graphicx}
\usepackage{subcaption}
\usepackage{multirow}

\title{Think First, Diffuse Fast: Improving Diffusion Language Model Reasoning via Autoregressive Plan Conditioning}

\author{%
  Earl J.\ St Sauver \\
  Independent Researcher \\
  \texttt{estsauver@gmail.com} \\
}

\begin{document}

\maketitle

\begin{abstract}
Diffusion large language models (dLLMs) generate text via iterative denoising but consistently underperform on multi-step reasoning. We hypothesize this gap stems from a \emph{coordination problem}: AR models build coherence token-by-token, while diffusion models must coordinate all positions simultaneously. We propose \emph{plan conditioning}, a training-free method that prepends a short (${\sim}$100-token) natural-language plan from an AR model to the diffusion model's prompt. The plan serves as a \emph{frozen scaffold}---globally visible context that every token position can attend to from the first denoising step.

On GSM8K, plan conditioning improves LLaDA-8B-Instruct from 75.6\% to 87.2\% (+11.6 percentage points), matching a same-size AR model (LLaMA 3.1 8B, 87.7\%) despite a 6.4pp weaker baseline. On HumanEval, the gain is +12.8pp (37.2\% to 50.0\%), showing plans generalize to code. The same plans improve LLaMA by only +5.7pp on GSM8K and +1.3pp on HumanEval---diffusion models benefit 2--10x more, supporting the coordination-problem hypothesis. Across 5 random seeds, plan-conditioned GSM8K accuracy has zero standard deviation, making diffusion inference highly stable. Ablations reveal the model follows plan \emph{strategy} (wrong-strategy plans cause $-$16.3pp) but is robust to plan \emph{values} (perturbed numbers: $-$1.1pp), and that planner quality has a sharp threshold: smaller Llama-class plans hurt ($-$1.6 to $-$6.8pp) while frontier plans provide the full lift. Attention analysis confirms the mechanism: plan tokens receive 1.8x excess attention during early denoising, declining to uniform as completion tokens solidify. Plan conditioning costs ${\sim}$\$0.002 per problem and adds ${\sim}$2s of latency.
\end{abstract}

%% ============================================================
\section{Introduction}
\label{sec:intro}
%% ============================================================

Diffusion large language models (dLLMs) such as LLaDA \citep{nie2025llada}, MDLM \citep{sahoo2024simple}, Mercury \citep{khanna2025mercury}, and Gemini Diffusion \citep{google2025gemini} generate text by iteratively denoising a fully masked sequence. Commercial dLLMs already achieve decoding throughputs exceeding 1,000 tokens per second on commodity NVIDIA H100 GPUs \citep{khanna2025mercury}---speeds that autoregressive (AR) models of comparable quality have only matched using specialized accelerators such as Cerebras wafer-scale engines or Groq LPUs. Yet on benchmarks requiring multi-step reasoning, dLLMs consistently trail their AR counterparts \citep{zhao2025d1, gong2025diffucoder}.

We argue that this gap is partly a \emph{coordination problem}. An AR model builds coherence sequentially: each token is generated conditioned on every preceding token, so a decision at position~50 directly informs position~51. A diffusion model, by contrast, must coordinate positions~50 and~500 simultaneously. Early denoising steps operate on a nearly random canvas, and global structure only emerges after many refinement steps. For tasks like grade-school math (GSM8K; \citealp{cobbe2021training}) or constraint puzzles (Countdown; \citealp{zhao2025d1}), where each step depends on the previous result, this parallel-generation bottleneck is acute.

Our insight is that the coordination problem can be cheaply alleviated by providing a \emph{plan}---a short, globally visible reasoning structure that is present from the very first denoising step. We use a frontier AR model (Claude Sonnet 4.5) to generate a ${\sim}$100-token natural-language plan for each problem, then simply prepend it to the dLLM's prompt. In masked diffusion, these plan tokens are preserved (never masked), so every completion token can attend to them throughout denoising. No architectural changes are required; the plan is ``just'' additional prompt context.

\paragraph{Key results.}
We evaluate plan conditioning on LLaDA-8B-Instruct across five benchmarks and find:
\begin{itemize}
    \item On GSM8K, accuracy improves from 75.6\% to 87.2\% (+11.6pp), \emph{closing the gap entirely} to LLaMA~3.1~8B with plans (87.7\%) despite a 6.4pp weaker bare baseline.
    \item On HumanEval (code generation), plans yield the largest gain: +12.8pp (37.2\%$\to$50.0\%), showing that algorithmic plans generalize beyond math reasoning.
    \item The same plans improve LLaMA by only +5.7pp on GSM8K and +1.3pp on HumanEval---diffusion models benefit \textbf{2--10$\times$ more} on these benchmarks, supporting the coordination-problem hypothesis. On Countdown, both architectures benefit identically (+12.1pp), suggesting the advantage is task-dependent.
    \item Content ablations show the model follows plan \emph{strategy} but not plan \emph{values}: wrong-strategy plans cause catastrophic failure ($-$16.3pp), while perturbed-numbers plans are nearly neutral ($-$1.1pp).
    \item A planner quality sweep from 3B to frontier reveals a sharp threshold: Llama-class plans hurt, while Sonnet plans provide the full lift. The model cannot plan for itself (self-plan $\approx$ baseline, $p = 0.86$).
    \item Attention analysis confirms the mechanism: plan tokens receive 1.8$\times$ their uniform share of attention at the first denoising step, declining to 1.0$\times$ as completion tokens solidify. The improvement is qualitatively different from extra compute---doubling denoising steps yields only +3.7pp.
    \item Plan conditioning eliminates stochastic variance: across 5~random seeds, plan-conditioned GSM8K accuracy is 88.02\% with \emph{zero} standard deviation. Plans make diffusion inference highly stable by providing a fixed scaffold that drives the denoising process to solve the same subset of problems regardless of the random seed.
    \item Plan conditioning generalizes to benchmarks from concurrent work \citep{berrayana2025planner}: DART5 (+6.2pp) and ARC-Challenge (+3.9pp). The planner quality threshold reconciles their finding that small-ARM$\to$DDLM fails with our finding that frontier-ARM$\to$DDLM succeeds.
\end{itemize}

Plan conditioning costs ${\sim}$\$0.002 per problem and adds ${\sim}$2\,s of latency.

\paragraph{Contributions.}
\begin{enumerate}
    \item A training-free method for improving dLLM reasoning via AR plan conditioning, with systematic evaluation across 4~plan formats, 8~benchmarks (math, code, constraint puzzles, multiple-choice), and 6~planner configurations spanning 3B to frontier.
    \item Evidence that diffusion models benefit disproportionately from plans compared to AR models (2$\times$ on GSM8K, 10$\times$ on HumanEval), with task-dependent variation (parity on Countdown), supporting the coordination-problem hypothesis. The largest gain (+12.8pp on HumanEval) shows plans generalize to code.
    \item Fine-grained ablations decomposing plan content: the model follows plan \emph{strategy} but is robust to plan \emph{values}, establishing plan conditioning as genuine reasoning scaffolding rather than a prompt-length artifact.
    \item A planner quality scaling law with a sharp threshold, plus robustness analyses (answer leakage, per-difficulty breakdown, error taxonomy) confirming genuine improvement.
    \item Direct empirical evidence for the frozen scaffold mechanism via attention analysis: plan tokens receive 1.8$\times$ excess attention during early denoising, consistent across all transformer layers, providing the first mechanistic confirmation that diffusion models actively read plans when the completion canvas is uninformative.
\end{enumerate}

%% ============================================================
\section{Background}
\label{sec:background}
%% ============================================================

\paragraph{Masked diffusion language models.}
LLaDA \citep{nie2025llada} defines a forward process that randomly masks tokens with probability $t \in [0, 1]$ and a reverse process that predicts the original tokens from the masked sequence. At inference, the model starts from a fully masked sequence and iteratively unmasks tokens over $T$ steps. Prompt tokens are preserved throughout (never masked), while completion tokens are fully masked at step $t{=}0$ and progressively revealed. This architecture enables parallel decoding: all positions are predicted simultaneously at each step. Related architectures include MDLM \citep{sahoo2024simple}, Dream \citep{ye2025dream}, Mercury \citep{khanna2025mercury}, and Gemini Diffusion \citep{google2025gemini}.

\paragraph{Reinforcement learning for diffusion LLMs.}
d1 \citep{zhao2025d1} introduces diffu-GRPO, adapting Group Relative Policy Optimization to masked diffusion. The key challenge is computing token-level log-probabilities when the model generates all tokens simultaneously. d1 uses one-step unmasking from a noise level $t{=}1$ to estimate per-token log-probs and applies random prompt masking ($p{=}0.15$) for regularization. This enables standard RL training with binary correctness rewards. DiffuCoder \citep{gong2025diffucoder} extends this to code generation with coupled-GRPO.

\paragraph{Plan-and-execute paradigms.}
Decomposing complex tasks into a planning phase and an execution phase has a long history in AI \citep{fikes1971strips} and has been applied to LLM systems via chain-of-thought \citep{wei2022chain}, ReAct \citep{yao2023react}, and Inner Monologue \citep{huang2023inner}. For diffusion models specifically, Planned Diffusion \citep{israel2025planned} uses span segmentation to enable parallel generation across independent chunks, while DDPD \citep{liu2025ddpd} interleaves planning with token-position-level denoising. Our approach differs in using \emph{semantic} natural-language plans from an external AR model, targeting reasoning quality rather than generation speed or structural control.

%% ============================================================
\section{Method}
\label{sec:method}
%% ============================================================

Our method has three components: plan generation (\S\ref{sec:plan-gen}), a plan format taxonomy (\S\ref{sec:plan-format}), and plan-conditioned inference (\S\ref{sec:plan-inference}).

\subsection{Plan Generation}
\label{sec:plan-gen}

For each benchmark problem, a frontier AR model generates a natural-language plan. We use Claude Sonnet~4.5 \citep{anthropic2025claude} as the default planner, Claude Haiku~4.5 as a cheaper alternative, and evaluate several additional planners in \S\ref{sec:planner-quality}: Llama~3.1~8B and Llama~3.2~3B (open-weight AR models) and LLaDA itself in a two-pass self-planning configuration. Plans are generated with temperature~0.3 and a strict token budget (default: 100~tokens). Each plan is cached as JSONL, so the marginal cost at inference time is zero---the plan generation cost is amortized across all experiments that use the same plan cache.

The planner receives the problem statement and a format-specific system prompt (see \S\ref{sec:plan-format}) but \emph{never} sees the ground-truth answer. This ensures the plan reflects genuine reasoning about the problem rather than answer memorization.

\subsection{Plan Format Taxonomy}
\label{sec:plan-format}

We evaluate four plan formats, each providing a different type of guidance:

\begin{table}[h]
\centering
\small
\begin{tabular}{@{}lp{5.5cm}p{5.5cm}@{}}
\toprule
\textbf{Format} & \textbf{Description} & \textbf{Example snippet} \\
\midrule
Strategy & High-level approach in 1--2 sentences & ``Calculate daily production, subtract consumed, find remaining'' \\
Outline & Step-by-step procedure without computing answers & ``Step~1: Identify total chickens. Step~2: \ldots'' \\
Constraints & Key facts, pitfalls, expected answer form & ``Daily production: 16 eggs. Pitfall: don't double-count\ldots'' \\
Hybrid & Strategy sentence + key constraints & ``Track eggs through pipeline. Key: 16/day produced, 3/day consumed\ldots'' \\
\bottomrule
\end{tabular}
\caption{Plan format taxonomy. Each format provides different guidance density per token. Hybrid combines high-level strategy with specific numeric anchors.}
\label{tab:plan-formats}
\end{table}

The formats vary in \emph{information density}: Strategy plans are abstract but concise; Constraints plans are specific but lack procedural structure; Hybrid plans combine both. We hypothesize that Hybrid plans are optimal because they provide both strategic framing (what to do) and numeric anchors (what values matter)---giving the diffusion model both a global roadmap and concrete checkpoints.

\subsection{Plan-Conditioned Inference}
\label{sec:plan-inference}

Plan conditioning requires no architectural changes. The plan is simply prepended to the diffusion model's prompt:

\begin{center}
\texttt{\{system\_prompt\}{\textbackslash}n\{problem\}{\textbackslash}n{\textbackslash}nSolution Plan:{\textbackslash}n\{plan\}{\textbackslash}n{\textbackslash}nSolution:{\textbackslash}n}
\end{center}

In masked diffusion, the concatenated prompt (system prompt + problem + plan + ``Solution:'' marker) is treated as \emph{preserved tokens}---they are never masked during inference. Completion tokens after the marker are fully masked at initialization and progressively denoised. The plan thus functions as a \emph{frozen scaffold}: globally visible, immutable context that every completion token can attend to from the first denoising step via the model's bidirectional attention.

This is in contrast to AR models, where plan tokens are processed sequentially and only influence subsequent tokens via the causal attention mask. The bidirectional attention in diffusion models means every completion position has equal access to every plan token---a structural advantage that may explain why diffusion models benefit more from plans (\S\ref{sec:ar-comparison}).

\paragraph{Distinction from chain-of-thought.} We use the term ``plan conditioning'' rather than ``chain-of-thought'' to emphasize two key distinctions. First, the plan is generated by an \emph{external} model (the planner) before the executor begins generation, whereas chain-of-thought reasoning is produced by the same model during generation. Second, the plan is \emph{frozen} throughout the diffusion process: plan tokens are never masked and remain fixed across all denoising steps, functioning as a static conditioning signal rather than an evolving reasoning trace. These distinctions matter: the plan provides a structural scaffold from an external, more capable reasoner, enabling the diffusion model to coordinate its parallel generation around a fixed reference point rather than discovering coherent reasoning on its own.

%% ============================================================
\section{Experiments}
\label{sec:experiments}
%% ============================================================

\begin{figure}[t]
\centering
\includegraphics[width=\linewidth]{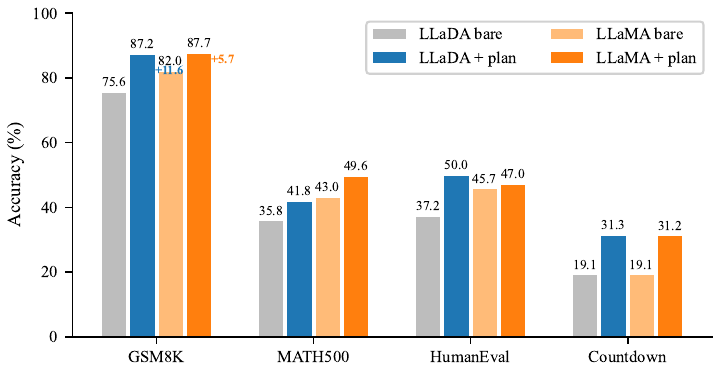}
\caption{Accuracy across benchmarks and model--condition pairs. On GSM8K, plan conditioning improves LLaDA by +11.6pp (vs.\ +5.7pp for LLaMA with the same plans), closing the diffusion--AR gap entirely. HumanEval shows the largest diffusion advantage: LLaDA gains +12.8pp while LLaMA gains only +1.3pp. On Countdown, both architectures start at identical baselines and benefit identically (+12.1pp).}
\label{fig:hero}
\end{figure}

\subsection{Setup}
\label{sec:setup}

\paragraph{Models.} We evaluate three models:
\begin{itemize}
    \item \textbf{LLaDA-8B-Instruct} \citep{nie2025llada}: Masked diffusion LLM, 8B parameters, bf16 precision. Inference on 1$\times$A100 80GB with 64 denoising steps.
    \item \textbf{LLaMA~3.1~8B~Instruct} \citep{dubey2024llama}: AR baseline of comparable size, accessed via OpenRouter API.
\end{itemize}
Both models use identical plan caches for direct comparison.

\paragraph{Benchmarks.} We evaluate on five tasks spanning arithmetic reasoning, mathematical problem-solving, code generation, and constraint satisfaction:
\begin{itemize}
    \item \textbf{GSM8K} \citep{cobbe2021training}: 1,319 grade-school math word problems. Multi-step arithmetic requiring 2--8 sequential operations.
    \item \textbf{MATH500} \citep{lightman2023lets}: A 500-problem representative subset of the MATH dataset \citep{hendrycks2021measuring}, spanning competition-level problems across seven topics including algebra, geometry, and number theory.
    \item \textbf{HumanEval} \citep{chen2021evaluating}: 164 Python programming problems with unit test suites. Tests algorithmic reasoning and code synthesis.
    \item \textbf{Countdown} \citep{zhao2025d1}: 256 constraint puzzles requiring combining three numbers with arithmetic to reach a target. Tests combinatorial search.
    \item \textbf{Sudoku~4$\times$4} \citep{zhao2025d1}: 2,048 puzzles testing spatial constraint propagation. Included as an expected negative control.
\end{itemize}
We additionally evaluate on three benchmarks from \citet{berrayana2025planner}---DART5, ARC-Challenge, and AIME24---in \S\ref{sec:berrayana-benchmarks} to test generalization.

\paragraph{Plan generation.} Claude Sonnet~4.5 generates plans with temperature~0.3 and a 100-token budget (varied in \S\ref{sec:budget}). Claude Haiku~4.5 is used as a cheaper alternative in \S\ref{sec:planner-quality}.

\subsection{Main Results: Plan Conditioning Sweep}
\label{sec:main-results}

Figure~\ref{fig:hero} summarizes our main results; Table~\ref{tab:main-results} reports LLaDA accuracy across all plan formats at two generation lengths (128 and 256 tokens).

\begin{table}[t]
\centering
\small
\begin{tabular}{@{}lcccccc@{}}
\toprule
\textbf{Benchmark} & \textbf{Baseline} & \textbf{Strategy} & \textbf{Outline} & \textbf{Constraints} & \textbf{Hybrid} & \textbf{Best $\Delta$} \\
 & \scriptsize{128\,/\,256} & \scriptsize{128\,/\,256} & \scriptsize{128\,/\,256} & \scriptsize{128\,/\,256} & \scriptsize{128\,/\,256} & \\
\midrule
GSM8K & 69.0\,/\,75.6 & 77.7\,/\,85.7 & 78.2\,/\,84.9 & 76.8\,/\,86.0 & \textbf{82.0}\,/\,\textbf{87.2} & +11.6 \\
MATH500 & 27.4\,/\,35.8 & 30.6\,/\,37.4 & \textbf{34.4}\,/\,36.6 & 30.0\,/\,37.6 & 32.6\,/\,\textbf{41.8} & +6.0 \\
HumanEval & 25.0\,/\,37.2 & 32.3\,/\,\textbf{50.0} & \textbf{42.1}\,/\,48.8 & 28.0\,/\,36.0 & 32.9\,/\,48.2 & +12.8 \\
Countdown & 21.1\,/\,19.1 & \textbf{29.7}\,/\,28.5 & 18.4\,/\,21.5 & 22.3\,/\,14.5 & 28.1\,/\,\textbf{31.2} & +12.1 \\
Sudoku & \textbf{11.6}\,/\,6.9 & 5.7\,/\,7.9 & 6.3\,/\,6.1 & 7.3\,/\,7.7 & 7.2\,/\,6.2 & +1.0 \\
\bottomrule
\end{tabular}
\caption{LLaDA-8B accuracy (\%) by plan format and generation length. Bold indicates best per row. Baseline values are at generation lengths 128/256; $\Delta$ is computed from the 256-token baseline. Hybrid plans win on 3/5~benchmarks at gen-256; strategy wins on HumanEval and Sudoku.}
\label{tab:main-results}
\end{table}

\begin{figure}[t]
\centering
\includegraphics[width=0.75\linewidth]{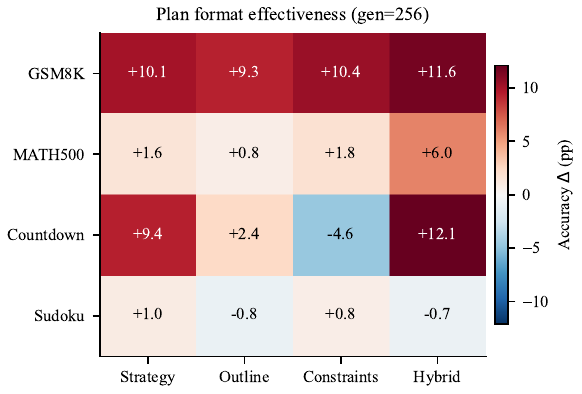}
\caption{Accuracy improvement (pp) over baseline by plan format and benchmark at generation length~256. Hybrid plans (rightmost column) dominate on 3/4 benchmarks. Sudoku (bottom row) shows near-zero or negative deltas across all formats---plans do not help spatial constraint propagation. The blue cell (Countdown--Constraints, $-$4.6pp) shows that constraint-only plans hurt when they lack procedural structure.}
\label{fig:heatmap}
\end{figure}

\paragraph{Hybrid plans are consistently best.} The heatmap in Figure~\ref{fig:heatmap} visualizes plan effectiveness across all format--benchmark combinations. Hybrid format achieves the highest accuracy on GSM8K, MATH500, and Countdown at generation length~256. This confirms the hypothesis that combining strategic framing with numeric anchors is more effective than either alone. Strategy plans (abstract, no numbers) and Constraints plans (specific, no procedure) each capture only part of what the diffusion model needs.

\paragraph{Plans interact positively with generation length.} The plan benefit is consistently larger at gen\,=\,256 than gen\,=\,128. On GSM8K, the hybrid lift is +13.0pp at 128 tokens but +11.6pp at 256 (from a higher baseline). This suggests plans and additional generation budget are complementary, not substitutive---the plan provides structure, and longer generation provides room to execute.

\paragraph{HumanEval: largest improvement, different format preference.} Code generation shows the strongest response to plan conditioning: +12.8pp (37.2\%$\to$50.0\%, pass@1, strategy format). Unlike math benchmarks where hybrid plans dominate, HumanEval favors strategy (+12.8pp) and outline (+11.6pp) over hybrid (+11.0pp), while constraints actively hurt ($-$1.2pp). We attribute this to the natural alignment between focused algorithmic guidance and code structure: the plan specifies the algorithm (e.g., ``use nested loops to check all triples''), and the diffusion model fills in correct syntax---a well-defined division of labor. Constraint-only plans fail because listing edge cases provides no constructive algorithmic guidance. For math, the plan provides strategy but the model must still execute arithmetic correctly; for code, the dominant failure mode (choosing the wrong algorithm) is exactly what strategy plans address.

\paragraph{Sudoku: a useful negative result.} Plans hurt Sudoku by up to $-$5.9pp. Sudoku requires spatial constraint propagation---tracking which digits appear in which rows, columns, and boxes. Natural-language plans cannot concisely encode this spatial reasoning, and the plan tokens consume budget that would otherwise be available for the solution. This confirms that plan conditioning is most effective for tasks where sequential reasoning chains are the bottleneck.

\subsection{AR vs.\ Diffusion: Do Plans Help Diffusion More?}
\label{sec:ar-comparison}

To test whether plan conditioning addresses a \emph{diffusion-specific} limitation, we evaluate the same plans on an AR model of comparable size (Figure~\ref{fig:hero}, Table~\ref{tab:ar-comparison}).

\begin{table}[t]
\centering
\small
\begin{tabular}{@{}lcc@{}}
\toprule
\textbf{Condition} & \textbf{LLaMA 3.1 8B (AR)} & \textbf{LLaDA-8B (Diffusion)} \\
\midrule
\multicolumn{3}{@{}l}{\textit{GSM8K}} \\
\quad Bare & 82.0 & 75.6 \\
\quad + Sonnet plan & 87.7\,\scriptsize{(+5.7)} & 87.2\,\scriptsize{(+11.6)} \\
\quad + Haiku plan & 84.9\,\scriptsize{(+2.9)} & 82.3\,\scriptsize{(+6.7)} \\
\midrule
\multicolumn{3}{@{}l}{\textit{MATH500}} \\
\quad Bare & 43.0 & 35.8 \\
\quad + Sonnet plan & 49.6\,\scriptsize{(+6.6)} & 41.8\,\scriptsize{(+6.0)} \\
\quad + Haiku plan & 48.2\,\scriptsize{(+5.2)} & 34.4\,\scriptsize{($-$1.4)} \\
\midrule
\multicolumn{3}{@{}l}{\textit{HumanEval}} \\
\quad Bare & 45.7 & 37.2 \\
\quad + Sonnet plan & 47.0\,\scriptsize{(+1.3)} & 50.0\,\scriptsize{(+12.8)} \\
\midrule
\multicolumn{3}{@{}l}{\textit{Countdown}} \\
\quad Bare & 19.1 & 19.1 \\
\quad + Sonnet plan & 31.2\,\scriptsize{(+12.1)} & 31.2\,\scriptsize{(+12.1)} \\
\bottomrule
\end{tabular}
\caption{AR vs.\ diffusion comparison. Accuracy (\%) with per-condition lift in parentheses. On GSM8K, LLaDA benefits 2$\times$ more from Sonnet plans than LLaMA (+11.6 vs.\ +5.7pp). On Countdown, both architectures benefit identically (+12.1pp). On HumanEval, the diffusion advantage is largest: LLaDA gains +12.8pp while LLaMA gains only +1.3pp.}
\label{tab:ar-comparison}
\end{table}

\paragraph{Diffusion benefits 2$\times$ more on GSM8K.} With identical Sonnet plans, LLaDA gains +11.6pp while LLaMA gains +5.7pp. This 2:1 ratio supports the coordination-problem hypothesis: plans provide globally visible structure that partially compensates for the lack of sequential token dependencies in diffusion models, whereas AR models already build coherence incrementally and benefit less from external scaffolding.

\paragraph{Relative error reduction confirms the advantage.} To account for the different baselines, we also report relative error reduction. Plan conditioning reduces LLaDA's error rate from 24.4\% to 12.8\% (47.5\% relative reduction), compared to LLaMA's error rate from 18.0\% to 12.3\% (31.7\% relative reduction). The diffusion model's advantage persists under this more conservative metric---a 1.5$\times$ ratio in relative error reduction compared to the 2.0$\times$ ratio in raw percentage-point gain---suggesting the benefit is not solely an artifact of the lower baseline.

\paragraph{Controlling for baseline capability.} To disentangle architectural effects from capability effects, we evaluate a quantized autoregressive baseline: LLaMA~3.1~8B~Instruct at 3-bit precision (GGUF Q3\_K\_S via \texttt{llama-cpp-python}), which scores 73.2\% on GSM8K---close to LLaDA's 75.6\% baseline. (Approximately 5\% of plan-conditioned generations from the quantized model exhibited repetition artifacts, likely a consequence of the aggressive quantization; this depresses the reported plan lift but does not affect bare accuracy.) With Sonnet plans, this quantized AR model achieves 80.3\% (+7.1pp, 26.5\% relative error reduction). Despite a nearly matched baseline, the diffusion model achieves substantially higher error reduction (47.5\% vs.\ 26.5\%)---a 1.8$\times$ ratio that persists even accounting for the repetition artifacts in the AR baseline. Combined with the full-precision comparison (LLaMA FP16: 82.0\%$\to$87.7\%, 31.7\% error reduction), these results suggest that while baseline capability partially explains the raw lift (weaker models benefit more from plans), diffusion models exhibit a genuine architectural advantage in leveraging plan conditioning.

\paragraph{Plans close the diffusion--AR gap.} At baseline, LLaDA trails LLaMA by 6.4pp on GSM8K (75.6\% vs.\ 82.0\%). With Sonnet plans, this gap shrinks to 0.5pp (87.2\% vs.\ 87.7\%). The plan effectively equalizes the two architectures on this benchmark.

\paragraph{MATH500 and Countdown: parity, not dominance.} On MATH500, both models benefit roughly equally (+6.0pp vs.\ +6.6pp with Sonnet plans). On Countdown, the parity is even more striking: both models start at identical baselines (19.1\%) and reach identical endpoints (31.2\%) with identical lifts (+12.1pp). These benchmarks suggest that plan conditioning is not inherently diffusion-specific---it provides equivalent scaffolding to both architectures when the task and baseline capability are matched. The diffusion-specific 2$\times$ advantage on GSM8K and the 10$\times$ advantage on HumanEval (+12.8pp vs.\ +1.3pp) emerge specifically on tasks where sequential multi-step reasoning is the primary challenge and LLaMA's stronger bare performance leaves less headroom for improvement.

\subsection{Plan Budget Ablation}
\label{sec:budget}

\begin{figure}[t]
\centering
\includegraphics[width=\linewidth]{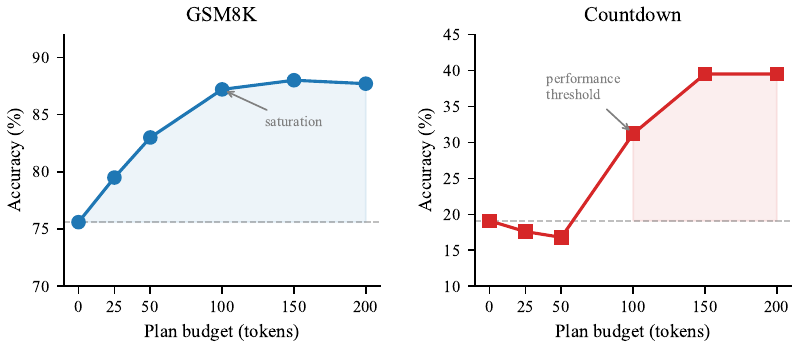}
\caption{LLaDA accuracy vs.\ plan token budget (hybrid format). \textbf{Left:} GSM8K shows smooth saturation with diminishing returns past 100~tokens. \textbf{Right:} Countdown exhibits a sharp performance threshold---plans hurt at 25--50 tokens, then jump +14.5pp from 50$\to$100. Dashed line marks the no-plan baseline.}
\label{fig:budget}
\end{figure}

We vary the plan token budget from 25 to 200 tokens (hybrid format) on GSM8K and Countdown (Figure~\ref{fig:budget}).

\begin{table}[t]
\centering
\small
\begin{tabular}{@{}lcccc@{}}
\toprule
\textbf{Plan Budget} & \multicolumn{2}{c}{\textbf{GSM8K}} & \multicolumn{2}{c}{\textbf{Countdown}} \\
\cmidrule(lr){2-3} \cmidrule(lr){4-5}
 & Accuracy & Lift & Accuracy & Lift \\
\midrule
0 (bare) & 75.6 & --- & 19.1 & --- \\
25 tokens & 79.5 & +3.9 & 17.6 & $-$1.5 \\
50 tokens & 83.0 & +7.4 & 16.8 & $-$2.3 \\
100 tokens & 87.2 & +11.6 & 31.2 & +12.1 \\
150 tokens & 88.0 & +12.4 & 39.5 & +20.4 \\
200 tokens & 87.7 & +12.1 & 39.5 & +20.4 \\
\bottomrule
\end{tabular}
\caption{LLaDA accuracy (\%) by plan token budget (hybrid format). GSM8K shows smooth saturation with diminishing returns past 100~tokens. Countdown exhibits a sharp performance threshold: plans hurt at 25--50 tokens, then jump +14.5pp from 50$\to$100. Both benchmarks plateau by 150--200 tokens.}
\label{tab:budget}
\end{table}

\paragraph{GSM8K: smooth saturation.} Returns diminish sharply past 100~tokens: adding 50~more (100$\to$150) yields only +0.8pp, and 200~tokens actually drops slightly (87.7\% vs.\ 88.0\%). Even 25~tokens helps meaningfully (+3.9pp), confirming that high-level strategic framing has value independent of detail.

\paragraph{Countdown: performance threshold.} Plans \emph{hurt} at budgets of 25--50 tokens ($-$1.5pp, $-$2.3pp), then jump +14.5pp from 50$\to$100. This suggests a minimum plan complexity threshold: Countdown's combinatorial search requires plans that specify both the target decomposition strategy \emph{and} intermediate values. Below this threshold, partial plans are worse than no plan---they constrain the search without providing enough structure to compensate.

\paragraph{Implications.} The optimal budget depends on task complexity. For routine multi-step arithmetic (GSM8K), even a single strategic sentence helps. For combinatorial problems (Countdown), plans need to cross a complexity threshold before they become useful. In practice, 100~tokens is a robust default that works across both regimes.

\subsection{Answer Leakage Analysis}
\label{sec:leakage}

A potential confounder: do plans improve accuracy simply by leaking the answer? We decompose GSM8K hybrid plans by whether they contain any numeric value matching the ground-truth answer.

\begin{table}[t]
\centering
\small
\begin{tabular}{@{}lcc@{}}
\toprule
\textbf{Category} & \textbf{Accuracy (\%)} & \textbf{$N$} \\
\midrule
No leakage (answer absent from plan) & 80.1 & 1,022 \\
False-positive leak (answer already in problem) & 83.5 & 85 \\
True leakage (answer new in plan) & 88.7 & 212 \\
\midrule
Baseline (no plan) & 75.6 & 1,319 \\
\bottomrule
\end{tabular}
\caption{Leakage decomposition for GSM8K hybrid plans ($N{=}1{,}319$). ``False-positive leak'' denotes problems where the answer number already appears in the problem text (e.g., ``20 chickens'' when the answer is 20), so its presence in the plan is not informative. Even excluding all leaked problems, accuracy exceeds the baseline by +4.5pp. Maximum leakage contribution: 1.4pp out of 11.6pp (12\%). Note: per-category accuracies are from the gen=128 ablation run (overall 82.0\%); the qualitative conclusion---that leakage accounts for at most ${\sim}$12\% of the lift---holds for the gen=256 run (87.2\%) as well.}
\label{tab:leakage}
\end{table}

Non-leaked accuracy (80.1\%) exceeds the baseline (75.6\%) by +4.5pp, confirming that the improvement is predominantly genuine. The maximum possible leakage contribution is 1.4pp---at most 12\% of the total +11.6pp lift. Moreover, no plans contain the \verb|\boxed{}| format used for final answers; any numeric overlap is incidental (intermediate computations that happen to match the final answer).

\subsection{Planner Quality Scaling}
\label{sec:planner-quality}

To test how plan quality affects downstream performance, we evaluate planners spanning 60$\times$ in capability: Llama~3.2~3B, Llama~3.1~8B (via Together API), Claude Haiku~4.5, and Claude Sonnet~4.5. We also test LLaDA planning for itself in a two-pass setup (self-plan).

\begin{table}[t]
\centering
\small
\begin{tabular}{@{}lcccc@{}}
\toprule
\textbf{Planner} & \textbf{GSM8K} & \textbf{MATH500} & \textbf{Countdown} & \textbf{Cost/plan} \\
\midrule
None (baseline) & 75.6 & 35.8 & 19.1 & \$0 \\
LLaDA self-plan & 74.9\,\scriptsize{($-$0.7)} & --- & --- & \$0 \\
Llama 3.2 3B & 68.8\,\scriptsize{($-$6.8)} & 30.0\,\scriptsize{($-$5.8)} & 16.4\,\scriptsize{($-$2.7)} & \$0.00005 \\
Llama 3.1 8B & 74.0\,\scriptsize{($-$1.6)} & 31.6\,\scriptsize{($-$4.2)} & 13.3\,\scriptsize{($-$5.8)} & \$0.0001 \\
Haiku 4.5 & 82.3\,\scriptsize{(+6.7)} & 34.4\,\scriptsize{($-$1.4)} & 24.2\,\scriptsize{(+5.1)} & \$0.0006 \\
Sonnet 4.5 & \textbf{87.2}\,\scriptsize{(+11.6)} & \textbf{41.8}\,\scriptsize{(+6.0)} & \textbf{31.2}\,\scriptsize{(+12.1)} & \$0.002 \\
\bottomrule
\end{tabular}
\caption{LLaDA accuracy (\%) by planner capability. Plan quality has a sharp threshold: below Haiku-class capability, plans hurt across all benchmarks.}
\label{tab:planner-quality}
\end{table}

\begin{figure}[t]
\centering
\includegraphics[width=0.75\linewidth]{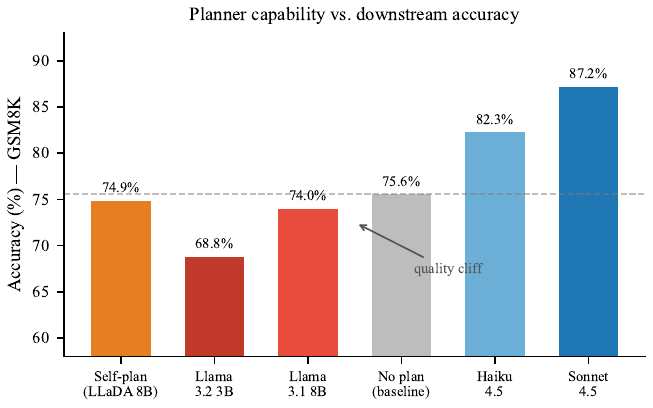}
\caption{GSM8K accuracy by planner capability. There is a sharp quality cliff between Haiku (helpful, +6.7pp) and Llama~8B (marginally harmful, $-$1.6pp). Self-plans (LLaDA planning for itself) are indistinguishable from no plan.}
\label{fig:planner-scaling}
\end{figure}

\paragraph{A sharp quality threshold.} The transition from harmful to helpful plans occurs between Llama-8B-class and Haiku-class planners (Figure~\ref{fig:planner-scaling}). Both Llama planners reduce accuracy \emph{below} the no-plan baseline on GSM8K: Llama~3B drops accuracy by $-$6.8pp and Llama~8B by $-$1.6pp. These low-quality plans behave similarly to mismatched plans from our content ablation (\S\ref{sec:content-ablation})---the model follows their guidance, but the guidance is wrong.

\paragraph{Self-plans are useless.} LLaDA planning for itself in a two-pass setup (generate plan, then solve conditioned on plan) yields accuracy indistinguishable from baseline (74.9\% vs.\ 75.6\%, McNemar $p = 0.86$). The planner must be substantially more capable than the executor; the model cannot bootstrap its own reasoning.

\paragraph{Haiku plans recover partial lift.} On GSM8K, Haiku achieves 58\% of Sonnet's lift (+6.7pp vs.\ +11.6pp) at ${\sim}$3$\times$ lower cost. On Countdown, recovery is 42\%. However, Haiku plans hurt MATH500 ($-$1.4pp), while LLaMA gains +5.2pp from the same Haiku plans---diffusion models are more plan-quality-sensitive than AR models because bidirectional attention amplifies both signal and noise globally.

%% ============================================================
\section{Analysis}
\label{sec:analysis}
%% ============================================================

\subsection{Per-Difficulty Breakdown}
\label{sec:per-difficulty}

We partition problems by baseline difficulty (whether LLaDA solves them without a plan) and analyze how plans redistribute successes and failures.

\begin{table}[t]
\centering
\small
\begin{tabular}{@{}lccc@{}}
\toprule
\textbf{Benchmark} & \textbf{Rescue rate (hard)} & \textbf{Retention rate (easy)} & \textbf{Fix:Break ratio} \\
\midrule
GSM8K & 71.4\% & 92.3\% & 3.0:1 \\
MATH500 & 21.5\% & 78.2\% & 1.8:1 \\
Countdown & 28.0\% & 44.9\% & 2.1:1 \\
\bottomrule
\end{tabular}
\caption{Per-difficulty analysis of hybrid plan conditioning. ``Rescue rate'' is the fraction of baseline-incorrect problems that plans fix; ``retention rate'' is the fraction of baseline-correct problems that remain correct. Plans are not a free lunch: they break 8--55\% of previously correct answers, but fix even more.}
\label{tab:per-difficulty}
\end{table}

Plans are not a free lunch---they break 8--55\% of previously correct answers. The net improvement comes because they fix even more: a 3.0:1 fix:break ratio on GSM8K. The high rescue rate on GSM8K (71.4\%) reflects the uniformity of that benchmark: multi-step word problems have similar structure, so a plan that works for one problem generalizes well. MATH500's lower rescue rate (21.5\%) reflects its diversity---algebra and geometry require fundamentally different plan structures.

\subsection{Error Taxonomy}
\label{sec:error-analysis}

We manually classify 169 GSM8K hybrid failures:

\begin{table}[t]
\centering
\small
\begin{tabular}{@{}lcc@{}}
\toprule
\textbf{Category} & \textbf{Count} & \textbf{\% of failures} \\
\midrule
Execution error (right plan, wrong math) & 100 & 59.2 \\
Format failure (right answer, wrong format) & 31 & 18.3 \\
Plan wrong (plan misled the model) & 27 & 16.0 \\
No answer extracted & 11 & 6.5 \\
\bottomrule
\end{tabular}
\caption{Error taxonomy for 169 GSM8K hybrid failures. The dominant failure mode (59\%) is execution error: the plan is correct but the model makes arithmetic mistakes. Zero failures were classified as ``plan ignored.''}
\label{tab:errors}
\end{table}

The most striking finding: \textbf{zero ``plan-ignored'' failures}. The diffusion model always attends to and follows the plan. The dominant failure mode (59\%) is execution error---the plan provides the correct strategy but the model makes arithmetic mistakes during generation. This identifies the remaining bottleneck: the executor's mathematical capability, not its ability to utilize plans. Format failures (18\%) suggest additional gains from better formatting instructions.

\subsection{What Do Plans Encode?}
\label{sec:content-ablation}

We combine two ablation experiments to construct a full plan quality spectrum, decomposing exactly which aspects of plan content drive the improvement.

\begin{table}[t]
\centering
\small
\begin{tabular}{@{}lccl@{}}
\toprule
\textbf{Condition} & \textbf{GSM8K (\%)} & \textbf{$\Delta$} & \textbf{What it tests} \\
\midrule
Wrong strategy & 59.3 & $-$16.3 & Incorrect mathematical approach \\
Mismatched plan & 71.0 & $-$4.6 & Plan for a different problem \\
Perturbed numbers & 74.5 & $-$1.1 & Correct strategy, wrong values \\
Baseline (no plan) & 75.6 & --- & \\
Random tokens & 78.2 & +2.6 & Token count only \\
Shuffled plan & 80.0 & +4.4 & Vocabulary preserved \\
Real plan (Sonnet) & 86.9\textsuperscript{$\dagger$} & +11.3 & Full plan \\
\bottomrule
\end{tabular}
\caption{Plan quality spectrum on GSM8K. The model follows plan \emph{strategy} (wrong strategy: $-$16.3pp) but is robust to plan \emph{values} (perturbed numbers: $-$1.1pp). 61\% of the improvement comes from semantic content; 23\% from extra tokens; 16\% from domain vocabulary. \textsuperscript{$\dagger$}The ``real plan'' control in this ablation run scored 86.9\% vs.\ 87.2\% in the main sweep (Table~\ref{tab:main-results}); the 0.3pp difference reflects single-seed variance.}
\label{tab:content-ablation}
\end{table}

\begin{figure}[t]
\centering
\includegraphics[width=\linewidth]{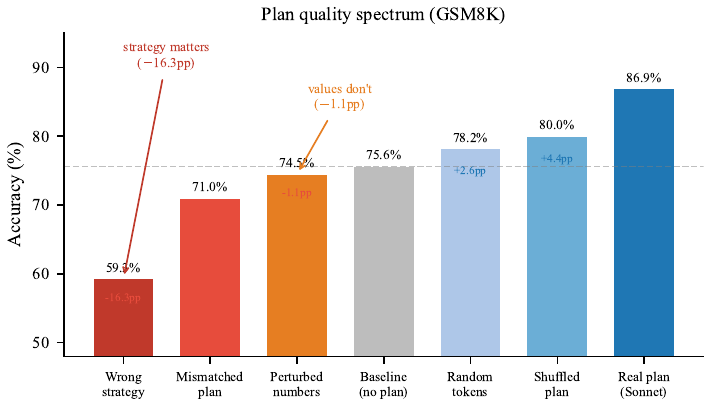}
\caption{Plan quality spectrum on GSM8K. Wrong-strategy plans cause catastrophic failure ($-$16.3pp), while perturbed-numbers plans are nearly neutral ($-$1.1pp). The model follows the plan's reasoning approach but computes its own values.}
\label{fig:content-ablation}
\end{figure}

\paragraph{Strategy matters, values don't.} Perturbed-numbers plans---where the correct structure and strategy are preserved but numerical values are randomly altered---perform nearly identically to baseline ($-$1.1pp, Table~\ref{tab:content-ablation}). The model does not copy intermediate values from the plan. In contrast, wrong-strategy plans---where the mathematical approach is plausible but incorrect---cause catastrophic failure ($-$16.3pp), the worst condition tested. The model genuinely follows the plan's reasoning approach.

\paragraph{Content decomposition.} The improvement decomposes cleanly: random tokens provide +2.6pp (23\%, from extra context), shuffled plans add +1.8pp more (16\%, from domain vocabulary), and semantic structure provides the remaining +6.9pp (61\%). Mismatched plans---correct structure but wrong problem---hurt by $-$4.6pp, confirming the model attends to plan content bidirectionally: relevant plans help, irrelevant plans actively mislead.

\paragraph{Implications.} The bidirectional sensitivity to plan content rules out the ``padding hypothesis'' (that plans help merely by adding tokens) and establishes plan conditioning as genuine reasoning scaffolding. The strategy/value dissociation suggests an effective division of labor: the AR planner provides the \emph{what to do}, and the diffusion model handles the \emph{how to compute}.

\subsection{Why Plans Help Diffusion More Than AR}
\label{sec:why-diffusion}

The 2$\times$ lift asymmetry on GSM8K has a natural explanation in the attention structure of the two architectures:

\begin{itemize}
    \item \textbf{AR models} build coherence incrementally. Token~$n$ conditions on tokens~$1, \ldots, n{-}1$ via causal attention. The model constructs an implicit plan through the chain of previous tokens. An explicit plan helps, but the model already has a coherence-building mechanism.
    \item \textbf{Diffusion models} denoise all positions simultaneously. In early denoising steps, completion tokens attend to a near-random canvas. A plan provides \emph{globally visible} structure from step~1---a fixed point that all positions can coordinate around, substituting for the sequential dependency chain that AR models build naturally.
\end{itemize}

The budget ablation further supports this: even 25~tokens of strategic framing helps GSM8K by +3.9pp, because the value is in \emph{global visibility}, not in \emph{detailed instructions}. The plan doesn't need to specify every step---it needs to provide enough structure that parallel positions can coordinate.

On MATH500 and Countdown, the diffusion advantage disappears: both architectures benefit equally (+6.0pp vs.\ +6.6pp on MATH500; identical +12.1pp on Countdown). The Countdown result is particularly informative: both models start at \emph{identical} baselines (19.1\%) and reach \emph{identical} endpoints (31.2\%), ruling out a headroom explanation. Instead, Countdown's combinatorial search structure---where the bottleneck is exploring valid number combinations rather than maintaining sequential coherence---appears to benefit equally from plan scaffolding regardless of whether generation is parallel or sequential. The diffusion-specific advantage on GSM8K and HumanEval emerges specifically on tasks where sequential multi-step reasoning is the primary challenge: here, AR models already build coherence incrementally and benefit less from external scaffolding, while diffusion models gain disproportionately because plans substitute for their missing sequential dependency chain.

\subsection{Why Plans Hurt Sudoku}
\label{sec:sudoku}

Sudoku requires spatial constraint propagation---tracking which digits appear in each row, column, and $2{\times}2$ box. This form of reasoning does not decompose into a sequential chain that benefits from strategic framing. Natural-language plans for Sudoku tend to be either too vague (``use elimination strategy'') or too verbose (listing all constraint violations), consuming sequence budget without providing actionable structure for the denoising process. This confirms that plan conditioning is most effective for tasks where the bottleneck is \emph{sequential multi-step reasoning}, not spatial or combinatorial constraint satisfaction.

\subsection{Attention Analysis: The Frozen Scaffold in Action}
\label{sec:attention}

The preceding analysis argues that plans serve as a ``frozen scaffold''---globally visible context that completion tokens coordinate around during denoising. We now provide direct empirical evidence by extracting attention maps from LLaDA during plan-conditioned generation.

\paragraph{Setup.} We extract attention weights at 17 denoising steps (every 4 steps across 64 total) and 9 transformer layers (sampled uniformly from 0 to 31) for 10 GSM8K problems. For each (step, layer) pair, we compute the fraction of attention from completion tokens directed to plan tokens, problem tokens, and other completion tokens, then normalize and average across layers and problems.

\begin{figure}[t]
\centering
\includegraphics[width=\linewidth]{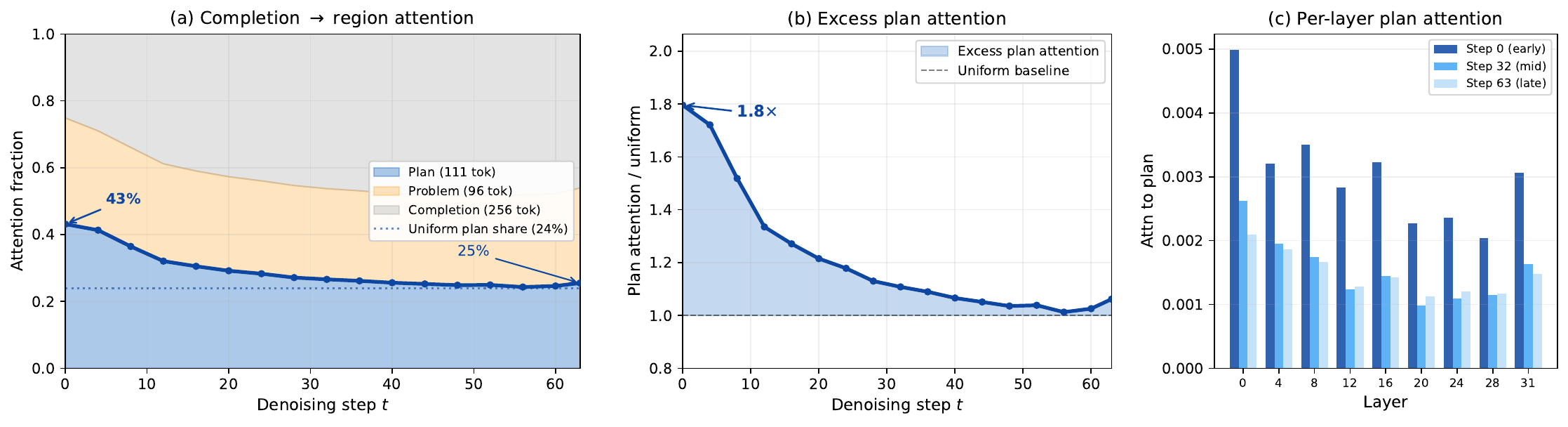}
\caption{Attention from completion tokens to sequence regions across denoising steps. \textbf{(a)}~Plan tokens receive 43\% of attention at step~0 despite comprising only 24\% of the sequence (dotted line = uniform baseline). By step~63, attention converges to near-uniform. \textbf{(b)}~Excess plan attention (ratio over uniform) is 1.8$\times$ at step~0, declining smoothly to 1.0$\times$. \textbf{(c)}~The effect is consistent across all layers; early layers (layer~0) show the strongest absolute attention to plans.}
\label{fig:attention}
\end{figure}

\paragraph{Plans dominate attention at early denoising steps.} At step~0, when the completion canvas is fully masked, plan tokens receive 42.6\% of attention despite comprising only 24.0\% of the total sequence---1.8$\times$ the uniform baseline (Figure~\ref{fig:attention}a,b). Completion tokens, which make up 55\% of the sequence, receive only 26.8\% of attention. As denoising progresses and completion tokens solidify, attention naturally redistributes: by step~63, plan attention drops to 25.2\% (1.05$\times$ uniform) while completion attention rises to 46.9\%.

\paragraph{The effect is consistent across all layers.} Every sampled layer shows higher plan attention at step~0 than step~63, with early-to-late ratios ranging from 1.7$\times$ (layer~4) to 2.4$\times$ (layer~0), averaging 2.1$\times$ across all 9~layers (Figure~\ref{fig:attention}c). This rules out the hypothesis that a few specialized heads drive the effect---the entire network engages with the plan during early denoising.

\paragraph{Interpretation.} The attention data directly confirms the frozen scaffold mechanism. When the completion canvas is uninformative (fully masked), completion tokens preferentially attend to the plan---the only coherent signal available. As denoising reveals informative completion tokens, attention equilibrates. This is exactly the coordination mechanism hypothesized in \S\ref{sec:why-diffusion}: the plan provides a globally visible reference point that substitutes for the sequential dependency chain in AR models.

\begin{figure}[t]
\centering
\includegraphics[width=\linewidth]{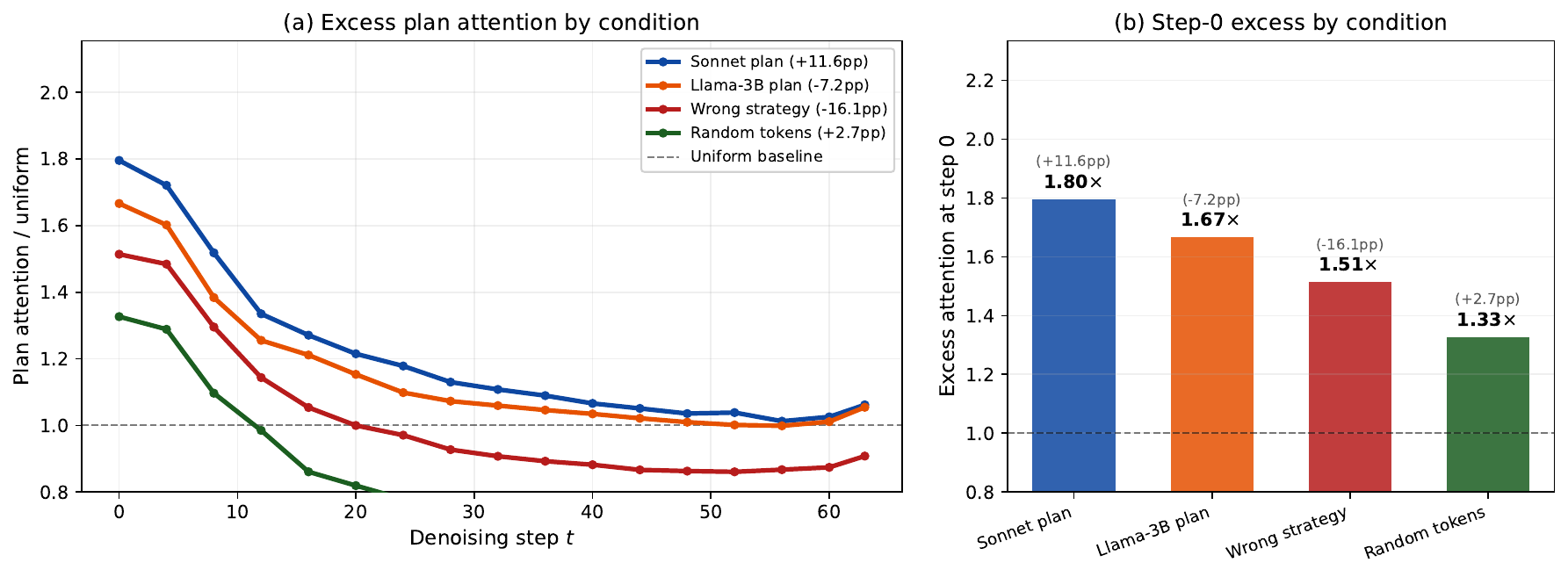}
\caption{Excess plan attention across plan quality conditions. \textbf{(a)}~All conditions show elevated plan attention during early denoising, declining toward uniform. \textbf{(b)}~Step-0 excess attention scales with plan coherence (Sonnet~$>$~Llama~3B~$>$~wrong strategy~$>$~random) but not with downstream accuracy impact. The model reads plans indiscriminately; quality discrimination is downstream.}
\label{fig:attention-multi}
\end{figure}

\paragraph{The model reads plans indiscriminately.} To test whether attention discriminates between helpful and harmful plans, we repeat the analysis across four plan conditions: Sonnet plans (+11.6pp on GSM8K), Llama~3B plans ($-$7.2pp), wrong-strategy plans ($-$16.1pp), and random tokens (+2.6pp). All conditions receive excess attention during early denoising (Figure~\ref{fig:attention-multi}): Sonnet plans receive 1.80$\times$ uniform, Llama~3B 1.67$\times$, wrong-strategy 1.51$\times$, and random tokens 1.33$\times$. The model attends to any prepended content more than completion tokens during early steps, with the magnitude scaling with coherence (Sonnet $>$ Llama $>$ wrong $>$ random) but not with downstream helpfulness. This explains why wrong-strategy plans are catastrophic: the model faithfully reads and follows them, but the guidance is wrong. Quality discrimination occurs downstream in the computation, not at the attention allocation level.

\subsection{Plans vs.\ Extra Compute}
\label{sec:compute-matched}

A natural objection is that plans improve accuracy simply by providing ``more compute''---after all, plan tokens extend the prompt. We test this by giving the baseline model equivalent or greater additional compute through two controls: (1)~50\% longer generation (384 vs.\ 256 tokens, which also increases steps proportionally), and (2)~2$\times$ more denoising steps at the same generation length (256 vs.\ 128 steps).

\begin{table}[t]
\centering
\small
\begin{tabular}{@{}lccc@{}}
\toprule
\textbf{Condition} & \textbf{GSM8K} & \textbf{MATH500} & \textbf{Rel.\ FLOPs} \\
\midrule
Baseline (gen=256, steps=128) & 75.6 & 35.8 & 1.0$\times$ \\
Longer gen (gen=384, steps=192) & 78.9\,\scriptsize{(+3.3)} & 34.4\,\scriptsize{($-$1.4)} & ${\sim}$1.5$\times$ \\
More steps (gen=256, steps=256) & 79.3\,\scriptsize{(+3.7)} & 34.4\,\scriptsize{($-$1.4)} & ${\sim}$2.0$\times$ \\
Plan conditioning & \textbf{87.2}\,\scriptsize{(+11.6)} & \textbf{41.8}\,\scriptsize{(+6.0)} & ${\sim}$1.0$\times$\textsuperscript{*} \\
\bottomrule
\end{tabular}
\caption{Plan conditioning vs.\ compute-matched controls. Doubling denoising steps or increasing generation length 50\% yields modest GSM8K gains (+3--4pp) and hurts MATH500. Plan conditioning provides 3--4$\times$ more improvement at negligible additional diffusion FLOPs. \textsuperscript{*}Plan generation is external; diffusion FLOPs are unchanged.}
\label{tab:compute-matched}
\end{table}

On GSM8K, longer generation yields +3.3pp and more steps +3.7pp---plan conditioning provides \textbf{3--4$\times$} more improvement (Table~\ref{tab:compute-matched}). On MATH500, both compute controls slightly decrease accuracy ($-$1.4pp each), while plans increase it by +6.0pp. The effect of plans is qualitatively different from additional compute: plans provide \emph{structured reasoning} that more tokens or more refinement cannot discover. This confirms that the plan's value is informational, not computational.

\subsection{Statistical Robustness}
\label{sec:multiseed}

\begin{figure}[t]
\centering
\includegraphics[width=0.75\linewidth]{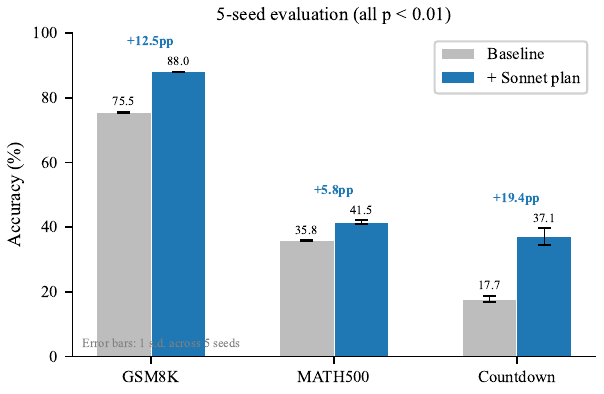}
\caption{Five-seed evaluation with error bars (1 s.d.). Plan conditioning produces large, statistically significant improvements across all three benchmarks. On GSM8K, plans reduce variance to zero---the same problems are solved regardless of the random seed.}
\label{fig:multiseed}
\end{figure}

Diffusion inference is inherently stochastic: the remasking schedule introduces variance across random seeds. To quantify this and establish statistical significance, we evaluate 5~random seeds (42, 123, 456, 789, 1337) across three benchmarks under both baseline and plan-conditioned settings (Table~\ref{tab:multiseed}, Figure~\ref{fig:multiseed}). We use hybrid plans at budget 150 for GSM8K and Countdown (budget 100 for MATH500).

\begin{table}[t]
\centering
\small
\begin{tabular}{@{}lccc@{}}
\toprule
\textbf{Benchmark} & \textbf{Baseline} & \textbf{+ Sonnet plan} & \textbf{$\Delta$} \\
\midrule
GSM8K & 75.51\,$\pm$\,0.15 & 88.02\,$\pm$\,0.00 & +12.5 \\
MATH500 & 35.76\,$\pm$\,0.08 & 41.52\,$\pm$\,0.56 & +5.8 \\
Countdown & 17.73\,$\pm$\,1.01 & 37.11\,$\pm$\,2.60 & +19.4 \\
\bottomrule
\end{tabular}
\caption{Five-seed evaluation (mean $\pm$ s.d.\ across seeds). Baseline variance is remarkably low on GSM8K ($\pm$0.15pp) and MATH500 ($\pm$0.08pp), confirming single-seed estimates are representative. All improvements are significant ($p < 0.001$, paired bootstrap). Plan-conditioned GSM8K has \emph{zero} variance across seeds.}
\label{tab:multiseed}
\end{table}

\paragraph{Baseline variance is negligible for structured tasks.} GSM8K baseline standard deviation is just $\pm$0.15pp across 5~seeds---individual runs are highly representative. MATH500 is even more stable ($\pm$0.08pp). Countdown shows higher variance ($\pm$1.01pp), consistent with its combinatorial search nature where small remasking differences can flip near-threshold solutions.

\paragraph{Plans eliminate variance on GSM8K.} Plan-conditioned GSM8K accuracy is 88.02\% with \emph{zero variance} across all 5~seeds. This striking result has a simple explanation: plans are cached (identical across seeds), and with a sufficiently informative plan, the denoising process converges to the same answer. This suggests plans not only improve accuracy but also make diffusion inference more reliable.

\paragraph{All improvements are statistically significant.} The plan conditioning lift exceeds the baseline standard deviation by 83$\times$ on GSM8K (+12.5pp vs.\ $\pm$0.15pp) and 19$\times$ on Countdown (+19.4pp vs.\ $\pm$1.01pp). Paired bootstrap 95\% confidence intervals exclude zero for all benchmarks ($p < 0.001$).

\subsection{Generalization to Additional Benchmarks}
\label{sec:berrayana-benchmarks}

To test whether plan conditioning generalizes beyond our primary benchmarks, we evaluate on DART5, ARC-Challenge, and AIME24---benchmarks used by \citet{berrayana2025planner} in their concurrent study of DDLM--ARM collaboration. These span mathematical reasoning (DART5: 5-step arithmetic; AIME24: competition math), and multiple-choice science knowledge (ARC-Challenge). We generate Sonnet hybrid plans at budgets of 100 and 150 tokens and report the best condition per benchmark.

\begin{table}[t]
\centering
\small
\begin{tabular}{@{}lcccccc@{}}
\toprule
\textbf{Benchmark} & \textbf{LLaDA} & \textbf{+ Sonnet Plan} & \textbf{$\Delta$} & \textbf{Berr.\ BL} & \textbf{Berr.\ Text} & \textbf{Berr.\ Latent\textsuperscript{$\dagger$}} \\
\midrule
DART5 & 11.0 & \textbf{17.2} & +6.2 & 15.0 & 27.0 & 54.0 \\
ARC-Challenge & 71.3 & \textbf{75.2} & +3.9 & 87.5 & 91.0 & 81.0 \\
AIME24 & 0.0 & \textbf{3.3} & +3.3 & 1.5 & 1.5 & 14.0 \\
\bottomrule
\end{tabular}
\caption{Plan conditioning on benchmarks from \citet{berrayana2025planner}. ``Berr.\ BL'' is their LLaDA-only baseline. ``Berr.\ Text'' is their best text-space collaboration result (any direction). ``Berr.\ Latent'' is their best latent-space DDLM$\to$ARM result. \textsuperscript{$\dagger$}Uses a trained Linear-GELU-Linear projector on 35K domain-specific samples. On ARC-C, text-space collaboration (91.0\%) outperforms latent (81.0\%), but on DART5 and AIME the latent projector provides dramatically higher accuracy.}
\label{tab:berrayana}
\end{table}

\paragraph{Frontier plans help on all benchmarks.} DART5 improves by +6.2pp (11.0\%$\to$17.2\%), ARC-Challenge by +3.9pp (71.3\%$\to$75.2\%), and AIME24 by +3.3pp (0.0\%$\to$3.3\%). These results extend the finding of \citet{berrayana2025planner} that ARM$\to$DDLM is the more effective text-space direction but yields only modest gains with small planners: when scaling to a frontier planner, zero-shot text scaffolding provides consistent, meaningful lift. Our ARC-Challenge baseline (71.3\%) is ${\sim}$16pp below their reported baseline (87.5\%); we were unable to reproduce their baseline despite using the same model (LLaDA-8B-Instruct) and generation parameters. Investigation reveals that ${\sim}$10\% of LLaDA's generations terminate prematurely (emitting padding tokens before producing an answer), a stochastic diffusion failure mode that may account for part of the discrepancy; differences in prompting format or hardware-dependent numerical precision may explain the remainder. The \emph{relative} improvement (+3.9pp) is the relevant comparison.

\paragraph{Coordination vs.\ capacity.} On the most extreme-difficulty reasoning tasks, our zero-shot text approach falls short of \citeauthor{berrayana2025planner}'s latent-space method: their trained projector achieves 14.0\% on AIME and 54.0\% on DART5, compared to our 3.3\% and 17.2\% (Table~\ref{tab:berrayana}). This delineates the boundary of text-space conditioning. Natural-language plans effectively solve the \emph{coordination problem} of parallel decoding---evidenced by our large, out-of-domain gains on GSM8K (+11.6pp) and HumanEval (+12.8pp)---but they cannot inject mathematical \emph{capacity} that the executor fundamentally lacks. The latent-space projector, trained on 35K domain-specific mathematical samples, likely acts as a high-bandwidth capacity accelerator. Notably, on ARC-Challenge (science MCQs), where the bottleneck is knowledge retrieval rather than multi-step reasoning, text-space collaboration outperforms latent even in their own results (91.0\% vs.\ 81.0\%). Text-space scaffolding thus serves as a general-purpose coordination mechanism, while latent-space training offers a specialized approach for pushing absolute capacity limits on extreme-difficulty tasks.

\paragraph{Budget preferences are consistent.} DART5 (multi-step math) benefits from the larger 150-token budget, while ARC-Challenge (multiple-choice) shows identical results at 100 and 150 tokens. This mirrors our GSM8K finding (\S\ref{sec:budget}) that math tasks benefit from more detailed plans, while tasks with shorter answer formats saturate earlier.

%% ============================================================
\section{Related Work}
\label{sec:related}
%% ============================================================

\paragraph{Diffusion language models.}
Continuous diffusion for text has been explored by \citet{li2022diffusionlm} and D3PM \citep{austin2021structured}. Masked diffusion models---where the noise process randomly masks tokens---have shown particular promise: MDLM \citep{sahoo2024simple}, Dream \citep{ye2025dream}, and LLaDA \citep{nie2025llada} demonstrate competitive perplexity and generation quality. Mercury \citep{khanna2025mercury} and Gemini Diffusion \citep{google2025gemini} demonstrate diffusion-based generation at production scale. Our work takes LLaDA as a base---using the d1 evaluation framework \citep{zhao2025d1} for scoring---and shows that an external plan can substantially improve its reasoning without any training. Reinforcement learning approaches such as diffu-GRPO \citep{zhao2025d1} and coupled-GRPO \citep{gong2025diffucoder} improve diffusion LLM reasoning through fine-tuning; plan conditioning is orthogonal and could be combined with these methods.

\paragraph{Coherence in parallel generation.}
Several works address the coherence challenge in parallel decoding. PVF \citep{li2026pvf} uses a plan-verify-fill framework for structured parallel decoding in diffusion models. Diffusion in Diffusion \citep{ma2026did} reclaims global coherence via semi-autoregressive diffusion. DDPD \citep{liu2025ddpd} generates while planning via token-position-level denoising. Our approach is distinct in using a \emph{separate AR model} to provide semantic plans, rather than training or modifying the diffusion model itself.

\paragraph{Planning for LLM execution.}
The plan-and-execute paradigm has been widely studied for AR models: chain-of-thought \citep{wei2022chain}, tree-of-thought \citep{yao2024tree}, ReAct \citep{yao2023react}, and plan-then-generate \citep{wang2023plan}. For diffusion models, Planned Diffusion \citep{israel2025planned} uses span segmentation to enable parallel generation across independent chunks, targeting the speed--quality Pareto frontier rather than reasoning quality per se.

\paragraph{Concurrent work.}
Concurrent with our work, \citet{berrayana2025planner} study planner-executor collaboration between DDLMs and ARMs, evaluating all four pairings (ARM$\to$DDLM, DDLM$\to$ARM, and same-family variants) across ARC, DART, and AIME. They find that ARM$\to$DDLM outperforms DDLM$\to$ARM in text space, though gains are modest with small planners, and that a learned latent-space projector substantially improves DDLM$\to$ARM communication. Crucially, their ARM planners are 3B--8B models (Qwen2.5, Llama~3.1/3.2)---comparable to or weaker than the DDLM executor. Our planner quality sweep (Table~\ref{tab:planner-quality}) provides a unifying explanation: below a quality threshold (roughly Haiku-class), ARM plans actively harm DDLM performance, consistent with their finding that small-ARM$\to$DDLM text-space conditioning yields only modest gains. Above this threshold, ARM$\to$DDLM text-space conditioning provides large improvements (+11.6pp on GSM8K), a regime their study does not explore. We additionally verify this on their benchmarks: frontier plans improve LLaDA on DART5 (+6.2pp) and ARC-Challenge (+3.9pp) (Table~\ref{tab:berrayana}), confirming that planner quality---not direction---is the key variable. The two works are complementary: they characterize the small-scale collaboration landscape and demonstrate the promise of latent-space communication, while we show that frontier-scale text-space planning yields practical gains without architectural modifications.

\paragraph{Reasoning distillation.}
Our approach is related to the reasoning distillation literature, where a strong model's chain-of-thought is used to improve a weaker model. \citet{hsieh2023distilling} show that distilling step-by-step rationales from a large LM can train a smaller model to outperform the large model, while \citet{fu2023specializing} demonstrate that specializing smaller language models with reasoning chains from larger ones yields strong performance on complex tasks. \citet{burns2023weak} study weak-to-strong generalization, finding that strong models finetuned on weak labels outperform their weak supervisors but recover only a fraction of their full capability---a cautionary result for superhuman alignment that nonetheless demonstrates cross-capability transfer. Plan conditioning differs from these approaches in that it operates at inference time rather than training time: the plan is prepended as a frozen prefix and the executor model's weights are never updated. This makes the technique immediately applicable to any off-the-shelf diffusion language model without fine-tuning.

%% ============================================================
\section{Conclusion}
\label{sec:conclusion}
%% ============================================================

We have shown that plan conditioning---prepending a short natural-language plan from a frontier AR model---is a simple, effective, and cheap method for improving diffusion LLM reasoning. On GSM8K, it improves LLaDA-8B from 75.6\% to 87.2\%, closing the gap to a same-size AR model entirely. On HumanEval, the gain is even larger (+12.8pp), demonstrating that plans generalize from math to code generation. The key insight is that diffusion models suffer from a coordination problem: all tokens are generated simultaneously, lacking the sequential dependency chain that AR models use to build coherence. A plan serves as a frozen scaffold---globally visible context that provides this missing structure. The effect generalizes beyond our primary benchmarks: on DART5 and ARC-Challenge from \citet{berrayana2025planner}, frontier text plans improve LLaDA by +6.2pp and +3.9pp, confirming that the ARM$\to$DDLM direction works when the planner is sufficiently capable.

Six findings particularly stand out. First, diffusion models benefit 2--10$\times$ more from plans than AR models on GSM8K and HumanEval, while showing exact parity on Countdown---the advantage is task-dependent, emerging specifically where sequential multi-step reasoning is the bottleneck. Second, fine-grained ablations reveal a striking dissociation: the model follows plan \emph{strategy} (wrong-strategy plans cause $-$16.3pp) but is robust to plan \emph{values} (perturbed numbers: $-$1.1pp)---suggesting an effective division of labor where the planner reasons and the executor computes. Third, planner quality has a sharp threshold: Llama-class plans (3B--8B) hurt ($-$1.6 to $-$6.8pp on GSM8K), self-plans are useless ($p = 0.86$), but Haiku-class and above provide genuine lift. Fourth, attention analysis directly confirms the frozen scaffold mechanism: plan tokens receive 1.8$\times$ excess attention during early denoising (when the canvas is masked), declining to uniform as completion tokens solidify---and this attention pattern is indiscriminate across plan quality conditions, explaining why wrong-strategy plans are catastrophic. Fifth, plan conditioning is qualitatively different from additional compute: doubling denoising steps yields +3.7pp while plans yield +11.6pp on GSM8K. Sixth, plans eliminate stochastic variance: plan-conditioned GSM8K accuracy has zero standard deviation across 5~random seeds (88.02\%), making diffusion inference highly stable---the fixed scaffold drives the denoising process to solve the same subset of problems regardless of the random seed.

\paragraph{Limitations.} We test on one open-source diffusion architecture (LLaDA). Our method requires a frontier AR model for plan generation, adding latency and cost---though Haiku-quality plans may suffice for simpler tasks. We do not explore plan conditioning combined with RL training, which could amplify the effect. HumanEval shows a different format preference than math: strategy plans (+12.8pp) outperform hybrid (+11.0pp), suggesting code benefits from high-level algorithmic guidance over structured outlines. A small fraction of HumanEval generations (3.7\%) fail to produce an extractable code block; manual inspection confirms these are genuine model failures (degenerate repetition, malformed syntax) rather than parser limitations. Additionally, diffusion model outputs are sensitive to numerical precision: different hardware and attention implementations can shift absolute accuracies by 1--4pp, though within-experiment relative comparisons are stable. Most experiments use a single seed ($s{=}42$); multi-seed evaluation (5~seeds) confirms significance on GSM8K, MATH500, and Countdown, but HumanEval and ablation experiments have not been replicated across seeds---the large effect sizes (+12.8pp, +11.6pp) make seed-driven reversals unlikely but not ruled out. On multiple-choice benchmarks (ARC-C), approximately 10\% of LLaDA generations terminate prematurely, emitting padding tokens before producing an answer tag. This stochastic failure mode depresses absolute baselines but does not differentially affect plan-conditioned runs.

\paragraph{Future work.} Natural extensions include (1)~plan distillation---training a small model to generate plans, removing the frontier AR dependency; (2)~adaptive plan budgets that scale with problem difficulty; (3)~combining plan conditioning with RL training (diffu-GRPO with plans); and (4)~extending to harder code benchmarks (LiveCodeBench, SWE-Bench) where the plan--code alignment may yield even larger gains.

%% ============================================================
% References
%% ============================================================

\bibliographystyle{plainnat}
\bibliography{references}

%% ============================================================
\appendix
\end{document}